\documentclass[letterpaper, 10 pt, conference]{ieeeconf}  

\IEEEoverridecommandlockouts                              

\usepackage{times}
\usepackage{soul}
\usepackage{url}
\usepackage[hidelinks]{hyperref}
\usepackage[utf8]{inputenc}
\usepackage[small]{caption}
\usepackage{subfigure}
\usepackage{graphicx}
\usepackage{amsmath,amssymb}
\usepackage{cite}

\usepackage{amsthm}
\usepackage{booktabs}

\usepackage{makecell}

\usepackage{algpseudocode,algorithm}
\usepackage{multirow}

\newcommand\iid{\mathrel{\overset{\makebox[0pt]{\mbox{\normalfont\tiny\sffamily i.i.d.}}}{\sim}}}
\newcommand{\MyInput}{\algorithmicrequire}
\renewcommand{\MyInput}{\textbf{Input:}}

\newcommand{\MyOutput}{\algorithmicrequire}
\renewcommand{\MyOutput}{\textbf{Output:}}

\DeclareMathOperator*{\argmin}{arg\,min}
\DeclareMathOperator*{\argmax}{arg\,max}

\title{\LARGE \bf
Adaptive t-Momentum-based Optimization for Unknown Ratio of Outliers in Amateur Data in Imitation Learning*
}

\author{Wendyam Eric Lionel Ilboudo$^{1}$,
        Taisuke Kobayashi$^{1}$,
        and~Kenji Sugimoto$^{1}$
\thanks{*This work was not supported by any organization}
\thanks{$^{1}$All authors are with the Division of Information Science, Nara Institute of Science and Technology, 8916-5 Takayama-cho, Ikoma, Nara 630-0192, Japan
e-mail: \{ilboudo.wendyam\_eric.in1, kobayashi, kenji\}@is.naist.jp}%
}

\begin{document}

\maketitle
\thispagestyle{empty}
\pagestyle{empty}

\begin{abstract}
Behavioral cloning (BC) bears a high potential for safe and direct transfer of human skills to robots. However, demonstrations performed by human operators often contain noise or imperfect behaviors that can affect the efficiency of the imitator if left unchecked. In order to allow the imitators to effectively learn from imperfect demonstrations, we propose to employ the robust t-momentum optimization algorithm. This algorithm builds on the Student's t-distribution in order to deal with heavy-tailed data and reduce the effect of outlying observations. We extend the t-momentum algorithm to allow for an adaptive and automatic robustness and show empirically how the algorithm can be used to produce robust BC imitators against datasets with unknown heaviness. Indeed, the imitators trained with the t-momentum-based Adam optimizers displayed robustness to imperfect demonstrations on two different manipulation tasks with different robots and revealed the capability to take advantage of the additional data while reducing the adverse effect of non-optimal behaviors.
\end{abstract}

\section{INTRODUCTION}

 The ultimate goal of the machine learning framework has always been to generate algorithms that perform at least as well as a human being, and robotics in particular, aims at building mechanical machines that can mimic human or animal behaviors. With this objective in mind, the Imitation learning (IL) approach has received an increasing attention, due to its ability to infer the hidden intention (policy) of an expert, which can be a human operator, through the observation of his/her demonstrations. In the literature, two types of IL are predominant: behavioral cloning (BC)~\cite{bain1995framework,torabi2018behavioral} which reproduces the sequences of the experts' action based on the environment state, and inverse reinforcement learning which maximizes a reward function inferred from the experts' demonstrations~\cite{ng2000algorithms,ho2016generative}. These algorithms have been shown to yield near-optimal policies when trained on high-quality demonstrations performed from experts, highlighting their potential for the production of advanced task-oriented robots that can naturally learn from demonstrations~\cite{dyrstad2018teaching,tsurumine2019generative}.

 Unfortunately, all these studies in both theoretical and applied aspects have assumed the presence of experts who always perform optimally, and of sophisticated operating interfaces that can adequately reflect the intentions of the experts, even when they do not make any mistakes. However, in practice, the demonstrators may lack qualitative expertise, either at the task itself or due to a non-intuitive operating interface, which means that they may be required to be well trained to become familiar with the setting before any demonstration can be recorded. However, this wastes both time and data, and constitutes an impractical constraint for crowdsourcing data collection~\cite{mandlekar2018roboturk}. Furthermore, even after being trained, a human operator may be subject to distractions due either to limited attention, tiredness or boredom, making the assumption of optimal and mistake-free demonstrations uncertain.
 For all these reasons, real world demonstrations are highly likely to contain unintentional noise and outliers, which makes it difficult for IL agents to extract an optimal policy.
Therefore, in general, such demonstrations which contains wrong actions would implicitly be excluded from the dataset used to train the agent, even when some parts of the demonstration may be informative.
Here, we define such a partially optimal demonstration as an \emph{amateur} demonstration.

 To tackle this issue and allow imitation from amateur demonstrations, several methods have been proposed. Indeed, for inverse reinforcement learning, we can cite the works~\cite{brown2019extrapolating} and~\cite{wu2019imitation} where additional labels provided by the experts are employed to discriminate amateur demonstrations, the work~\cite{tangkaratt2020variational} which assumes the amateur actions and states to be a Gaussian distributed noise, and the recent work~\cite{tangkaratt2020robust} where a pseudo-labeling technique is used to estimate the data density of the non-expert demonstrations and then a classification risk optimization is performed on all the demonstration dataset, using a symmetric loss function.

 In this study, we focus on the neural-network-based BC and, by seeing that amateur demonstrations include noise and outliers, employ the robust t-momentum~\cite{ilboudo2020tmomentum} optimization algorithm to train the imitator. With the t-momentum strategy, the adverse effect of noise and outliers can be implicitly removed according to its robustness hyperparameter during the stochastic gradient descent (SGD) updates. However, in the original version of the t-momentum, the robustness hyperparameter is needed to be specified before training and is therefore incapable of adapting automatically to the unknown actual ratio of noise and outliers inside amateur data. To address this issue, we extend the t-momentum with a method to automatically adjust the algorithm's robustness in order to deal with the uncertainty on the ratio of real world wrong demonstrations data for robotics application.

\section{PRELIMINARIES}

\subsection{Behavioral cloning}

 Behavioral cloning (BC)~\cite{bain1995framework} is an imitation learning technique which uses a supervised learning approach to capture and reproduce the behavior of a demonstrator, usually referred to as the \emph{expert}. As the expert performs the task, his/her actions are recorded along with the state that gave rise to the action. The sequence of these state-action records, called \emph{behavior trace} or \emph{trajectory}, is then used as supervised input signals for the \emph{imitator}, whose goal is to uncover a set of rules that reproduce the observed behavior. BC is powerful in the sense that the imitator is capable of immediately imitating the demonstrator without having to interact with the environment, making it particularly attractive for robotics applications and for safe and direct transfers of humans sub-cognitive skills or behaviors to machines.

 Formally, BC is concerned with the problem of finding a good imitation policy from a set of state-action demonstration trajectories $D_{\mathrm{demo}} = \{\tau_1, \tau_2, \cdots\}$ where $\tau$ is a trajectory $\{(s_1, a_1), \cdots, (s_N, a_N)\}$. This set of state-action pairs are used to seek the parameters $\theta$ of an imitation policy $\pi_{\theta}$ that best fits the set. This decision problem is usually solved by employing the maximum-likelihood estimation method. Indeed, assuming each $(s, a)$ pairs in $D_{\mathrm{demo}}$ are independently and identically distributed (i.i.d) and for $\pi_{\theta}$ defined as the imitator's policy parameterized by $\theta$, BC solves for an optimal solution $\theta^{*}$ such that:
\begin{align}
\theta^{*} &= \underset{\theta}{\argmax} \prod\limits_{(s_i, a_i) \sim D_{\mathrm{demo}}} \pi_{\theta}(a_i | s_i) \\
&= \underset{\theta}{\argmin} \sum\limits_{(s_i, a_i) \sim D_{\mathrm{demo}}} - \log \pi_{\theta}(a_i | s_i)
\label{eq:bc_optim}
\end{align}

 With this objective, the imitator's policy $\pi_{\theta}$ eventually converges to the unknown policy $\pi^{*}$ that produced the dataset $D_{\mathrm{demo}}$.

\subsection{Robust optimization with the t-momentum}

\subsubsection{Student's t-based momentum}

 Under the deep learning framework, complicated functions such as the policies $\pi_{\theta}(a | s)$ can be approximated using neural networks, where the parameters $\theta$ are given by the weights and biases of the networks.
 With the neural networks, the optimization problem depicted in Eq.~\eqref{eq:bc_optim} is solved by first-order gradient-based optimization methods.
 Most of the recent and popular first-order gradient-based methods developed nowadays build upon the momentum strategy~\cite{adam}, where an average of the past gradients are employed in the stochastic gradient descent updates.

 At the heart of the momentum methods' success lies the Exponential Moving Average (EMA), which allows recent gradients to have a greater impact on the average due to higher weights, while slowly forgetting observations that are far in the past and that possesses exponentially smaller weights. Let $J(\theta_t | S_t)$ be the objective function evaluated on a random sample $S_t$ from the training dataset, e.g. a sub-sample set of state-action pairs of size $m$ in BC $S_t = \{(s_i, a_i)\}_{i=1}^{m} \iid D_{\mathrm{demo}}$, and with the parameters $\theta_t$ at time $t$ corresponding to the weights and biases. With $g_t = \nabla_{\theta_t} J(\theta_t | S_t)$ the stochastic gradient of $J(\cdot)$ with respect to the parameters $\theta_t$, the regular EMA-based first-order momentum is defined as:
\begin{align}
    m_t &= \beta m_{t-1} + (1 - \beta) g_t \label{eq:adam_m}
\end{align}
 where $\beta \in (0, 1)$, the exponential decay coefficient, is a fixed value that controls how fast past gradients $g_i$, $i < t$, are forgotten.

 However, EMA-based momentum methods lack robustness to aberrant values due to the fact that every new observation is given the same weight $(1 - \beta)$.
 This led to the proposition of the t-EMA, a new EMA algorithm derived from the Student's t-distribution likelihood estimator, and its corresponding momentum, the \emph{t-momentum}~\cite{ilboudo2020tmomentum}. The particularity of the t-momentum lies in the fact that the decay coefficient $\beta_w$ is no longer fixed, but adaptive, and depends on the squared Mahalanobis distance $D_t$:
\begin{align}
    m_t = \beta_w m_{t-1} + (1 - \beta_w) g_t \label{eq:tadam_m}
\end{align}
where
\begin{align}
    \beta_w &= \frac{W_{t-1}}{W_{t-1} + w_t} \;,\;\;\mathrm{with}\; 
    w_t = \frac{\nu + d}{\nu + D_t} \label{eq:tadam_w} \\
    W_t &= \left(\frac{2\beta - 1}{\beta}\right) W_{t-1} + w_t \label{eq:tadam_W} \\
    D_t &= \sum_j^d \frac{(g_t^j - m_{t-1}^j)^2}{(\sigma_{t-1}^j)^2+\epsilon} \label{eq:tadam_mahalan_dist}
\end{align}
 where $\nu$ is the Student's t-distribution \emph{degrees of freedom} parameter which controls the robustness, $j$ in the superscript refers to the $j-$th component of the vector, and $(\sigma_{t-1})^2$ is an exponential moving variance estimate at step $t-1$, which is computed by default in recent methods.
 When integrated to momentum-based optimization methods such as Adam (Adaptive moment estimate)~\cite{adam}, the t-momentum has been shown to improve the robustness of the underlying optimizer and therefore increase the performance of the learning process against heavy-tailed data sets.

\subsubsection{t-EMA with modified weight decay}
 The decay strategy of the accumulated weights $W_t$ in Eq.~\eqref{eq:tadam_W} implies that at the time step $t$, the past value $w_{t-1}$ is not decayed with respect to the new value $w_t$ and that both have the same importance in the value of $\beta_w$.

 In order to ensure that the past value $w_{t-1}$ is decayed and has less importance than the new value $w_t$, Eq.~\eqref{eq:tadam_W} has been modified in~\cite{kobayashi2021tsoft} to yield instead:
\begin{align}
W_t &= \beta \left( W_{t-1} + w_t \right) \label{eq:tadam_modW}
\end{align}
which remains consistent with the maximum likelihood derivation of the t-momentum algorithm as described in~\cite{ilboudo2020tmomentum} and where the change of the decay factor's value, from $(2\beta - 1)/\beta$ in Eq.~\eqref{eq:tadam_W} to $\beta$, is set by the requirement that the t-EMA reverts to the EMA in the limit $\nu \rightarrow \infty$. With this modification, the value of $\beta_w$ at the time step $t$ is given by:
\begin{align}
\beta_w &= \frac{\beta W_{t-2} + \beta w_{t-1}}{\beta W_{t-2} + \beta w_{t-1} + w_t}
\end{align}
where the value of $w_{t-1}$ is effectively reduced with respect to $w_t$.

 In this study, this modified version of the t-EMA is the one we employ for the t-momentum.

\section{ROBUST BEHAVIORAL CLONING WITH ADAPTIVE T-MOMENTUM OPTIMIZATION}

\subsection{The imperfect demonstrations issue in behavioral cloning}

 Because BC relies solely on the provided demonstrations in order to find the imitation policy through a supervised learning approach, it requires all trajectories in the dataset to be optimal (i.e. perfect demonstrations) or near-optimal. Due to this fact, human operators, when given a control interface with the task to perform demonstrations, must first be trained to become highly efficient at using the interface before they can start demonstrating for the imitator; and even after having been trained, distractions, mistakes and limited attention time makes it difficult and nearly impossible for a human to always follow an optimal policy. This leads to trajectories where some state-action pairs are not optimal, causing the imitator to be biased against the optimal policy.

 We again refer to these imperfect demonstrations as being \emph{amateur} demonstrations, so that the dataset is generated as a mixture of the expert policy and the amateur policy:
\begin{align}
D_{\mathrm{demo}} \sim \rho(a, s) = (1 - \alpha) \rho_{\mathrm{exp}}(a, s) + \alpha \rho_{\mathrm{am}}(a, s)
\label{eq:am_exp_prop}
\end{align}
where $\rho_{\mathrm{exp}}$ and $\rho_{\mathrm{am}}$ are respectively the state-action density of the expert policy $\pi_{\mathrm{exp}}(a|s)$ and amateur policy $\pi_{\mathrm{am}}(a|s)$, $\alpha$ represents the proportion of amateur state-action pairs in the dataset, assumed to be in the range $[0.0, 0.5)$.

 In the original setting of behavioral cloning, all of the amateur demonstrations are simply discarded so that the policy that produced the dataset is only from the expert, i.e. $\pi^{*} = \pi_{\mathrm{expert}}$; however, this results in a loss of valuable data since all of the amateur $(s, a)$ pairs are not necessarily wrong. Due to the fact that BC typically require a lot of data in order to produce an optimal policy~\cite{ross2011reduction}, a strategy that takes advantage of good parts of the amateur demonstration (state-actions that are similar to the expert's one), while ignoring wrong or misleading actions in the imperfect demonstration is desirable.
 In this study, we propose to treat the amateur's imperfect demonstrations as being outliers and we show empirically how the t-momentum, a robust optimization algorithm, extended to allow adaptive robustness, can produce robust imitators in face of the resultant heavy-tailed dataset.

\subsection{Adaptive t-momentum for automatic robustness}

 The robustness of the Student's t-distribution, and therefore of the t-momentum derived from it, is controlled by the degrees of freedom parameter $\nu$. Indeed, as can be seen in Eq.~\ref{eq:tadam_w}, if $\nu \rightarrow \infty$, then $w_t \rightarrow 1$ for all time step $t$ and every values are given the same weight independently of the value of $D_t$, leading back to the non robust EMA derived from Gaussian distribution. In contrast, if $\nu \rightarrow 0$, then each value is weighted by $w_t = d/D_t$, leading to a strong sensitivity to the squared Mahalanobis distance and therefore to a very strong filtering effect. In the formulation of the t-momentum in the original t-momentum paper~\cite{ilboudo2020tmomentum}, the degrees of freedom is treated as an hyperparameter whose value must be set before starting the optimization process, meaning that the robustness of the t-momentum is fixed throughout the learning operations.

 In practice, the proportion value $\alpha$, introduced previously, is unknown (due to the difficulty to keep track of all imperfect $(s, a)$ pairs). Although one may analyze the dataset to infer its heaviness before starting training the imitator, in this section, a method for automatically adjusting the robustness of the t-momentum, based on the amount of outlying gradients encountered during training, is introduced.

 This mechanism exploits the batch approximation algorithm developed in~\cite{aeschliman2010novel}, in particular, the incremental version of the algorithm which is an efficient set of formulas capable of iteratively estimating the degrees of freedom for a given set of data points. Thanks to its incremental nature, the data do not need to be saved in memory and are instead treated sequentially as they are observed. This feature is of prime importance in the case of optimization methods, where the gradients are observed one at a time and can be arbitrarily large, rendering it difficult to store every one of them in memory. In the following, we refer to this algorithm as the Aeschliman's algorithm.

\subsubsection{Direct incremental degrees of freedom estimation algorithm}

 In order to compute an estimate for the degrees of freedom $\nu$, the Aeschliman's direct incremental algorithm is described as it follows: at each step $t$,
\begin{enumerate}
\item Compute a robust estimate for the mean $\mu_{t}$, such as the median.
\item Compute the logarithm of the squared euclidean norm of the difference between the recent observed data point $\boldsymbol{x}_t \in \mathbb{R}^d$ and the robust mean: $z_t = \log ||\boldsymbol{x}_t - \mu_{t}||^2$.
\item Update the arithmetic variance $\tilde{z}$ and mean $\bar{z}$ of the variable $z$:
\begin{align}
\tilde{z}_t &= \frac{1}{t}\tilde{z}_{t-1} + \frac{t-1}{t^2} \left(z_t - \bar{z}_{t-1}\right)^2 \label{eq:arithm_z_var}\\
\bar{z}_t &= \frac{1}{t}\bar{z}_{t-1} + \frac{t-1}{t}z_t
\label{eq:arithm_z_mean}
\end{align}
\item Compute a new estimate for the degrees of freedom:
\begin{align}
\nu_t &= \frac{1 + \sqrt{1 + 4b_t}}{b_t} \label{eq:aeshliman_dof}
\;,\;\;\mathrm{with}\; b_t = \tilde{z}_t - \psi_1 (\frac{d}{2})
\end{align}
where $\psi_1 (\cdot)$ is the trigamma function.
\end{enumerate}

\subsubsection{t-momentum with adaptive degrees of freedom}

 In order to integrate this algorithm to the t-momentum, a few changes are made to Aeschliman's algorithm, mainly in order to reduce the computational cost as much as possible. Namely,
\begin{itemize}
\item The t-momentum is directly used as the estimate of the robust mean, instead of computing the median as Aeschliman et al. did in their paper. Since the t-momentum is considered to be a robust mean estimate, this modification remains consistent with the original algorithm and it avoids the burden of estimating the gradient median, removing the need for a new variable.
\item Secondly, the squared norm in the variable $z$ computation is replaced by the squared Mahalanobis distance from equation~\eqref{eq:tadam_mahalan_dist}, i.e. $z_t = \log D_t$. This modification remains consistent with the original algorithm and can be understood as replacing the variable $\boldsymbol{x}_t$ by a standardized alternative $\frac{\boldsymbol{x}_t - \mu}{\sigma}$ who has $0$ mean and variance equals to $1$.
\item Finally, the arithmetic estimates for the variance and mean of the variable $z$ is replaced by exponential moving averages, i.e., the equations~\eqref{eq:arithm_z_var} and \eqref{eq:arithm_z_mean} becomes:
\begin{align}
\tilde{z}_t &= \lambda \tilde{z}_{t-1} + \lambda (1-\lambda) \left(z_t - \bar{z}_{t-1}\right)^2 \label{eq:ema_z_var}\\
\bar{z}_t &= \lambda\bar{z}_{t-1} + (1-\lambda) z_t
\label{eq:ema_z_mean}
\end{align}
 With $\lambda \in (0, 1)$. This particular modification is necessary in order to take into account the fact that machine learning tasks may be non-stationary, which requires the estimated mean and variance of $z$ to adapt to the changing data distribution.
\end{itemize}

 The new algorithm is named \emph{adaptive Student's t-distribution based momentum} or in short \emph{At-momentum} and the pseudo-algorithm is given in Algorithm~\ref{alg:atema}.

\begin{algorithm}[tb]
   \caption{At-momentum}
   \label{alg:atema}
   \MyInput{ Gradient $g_t$, Previous t-momentum: $m_{t-1}$, Previous weight's sum: $W_{t-1}$, Previous variance estimate: $\sigma^2_{t-1}$} \\
   \MyInput{ Previous mean and variance of $z$: $\tilde{z}_{t-1}$, $\bar{z}_{t-1}$}
   \begin{algorithmic}[1]
      \Require{$\beta$, $\lambda$: EMA decay parameters}
        \State{$d \gets dim[g_t]$}
        \State{$D_t \gets \sum_j^d \frac{(g_t^j - m_{t-1}^j)^2}{(\sigma_{t-1}^j)^2 + \epsilon}$}
        \State{$z_t \gets \log D_t$}
        \State{$\tilde{z}_t \gets \lambda \tilde{z}_{t-1} + \lambda (1-\lambda) \left(z_t - \bar{z}_{t-1}\right)^2$}
        \State{$\bar{z}_t \gets \lambda\bar{z}_{t-1} + (1-\lambda) z_t$}
        \State{$b_t \gets max(\epsilon,\;\tilde{z}_t - \psi_1 (\frac{d}{2}))$}
        \State{$\nu_t \gets d \left(1 + \sqrt{1 + 4b_t}\right) b_t^{-1}$}
        \State{$w_t \gets (\nu_t + d)\left (\nu_t + D_t \right )^{-1}$}
        \State{$\beta_w \gets W_{t-1} \left(W_{t-1} + w_t \right)^{-1}$}
        \State{$m_t \gets \beta_w m_{t-1} + (1-\beta_w) g_t$}
        \State{$W_t \gets \beta \left( W_{t-1} + w_t \right)$} \Comment{Decay the weights' sum}
   \end{algorithmic}
   \MyOutput{ $m_{t}$, $W_{t}$, $\tilde{z}_{t}$, $\bar{z}_{t}$}
\end{algorithm}

 Note that, for the practical implementation, the modified Aeschliman's algorithm is employed to estimate the degrees of freedom scale factor $k$, and the degrees of freedom is obtained using the equation $\nu_t = k_t \cdot dim[g_t]$ as suggested in the original t-momentum paper~\cite{ilboudo2020tmomentum}. This is necessary in order to keep the updates for being overly robust, since the Aeschliman's algorithm tends to produce small values for the degrees of freedom, which, when compared to the dimension of the neural network gradients can be negligible.

\section{EXPERIMENTS}

\subsection{Algorithm setup}

\subsubsection{Optimization algorithm's choice}

 In the following, we employ the t-Adam~\cite{ilboudo2020tmomentum} optimizer which is the Adam~\cite{adam} optimizer augmented with the t-momentum. The Adaptive t-momentum version is called \emph{At-Adam} and in order to investigate the effect of the decay parameter $\lambda$ used for the mean and variance of $z$ in equations~\eqref{eq:ema_z_mean} and~\eqref{eq:ema_z_var}, two values are defined:
\begin{itemize}
\item one that takes the same value as the considered momentum (here the first-order momentum of Adam) decay factor, i.e. $\lambda = \beta_1 = 0.9$, and
\item a larger value, which is set to be equal to the decay factor of the Adam second moment, i.e. $\lambda = \beta_2 = 0.999$.
\end{itemize}
 The results of training with Adam, without the t-momentum's robustness, are also included for reference.

\subsubsection{Policy model description}

 For all experiments, the imitator agent's policy model is implemented by a PyTorch~\cite{paszke2019pytorch} neural network with five hidden linear layers made of 100 neurons each, fits out with a layer normalization~\cite{ba2016layer} and with the ReLU activation function. The outputs are the actions' mean and covariance matrix diagonal elements for a multivariate Gaussian distribution.  Different random seeds are used for each models, but all optimizers share the same set of seeds, e.g. for $5$ trained models, the set is $\{ 1, \cdots, 5\}$.

\subsubsection{Performance measure}

 For all experiments, we run each of the trained models on the real robot for a certain number of times (most often 5 times), and count the number of times the imitator was capable of solving the given task. This performance measure is then represented by the \emph{success rate}:
\begin{align}
{\mathrm{success} \; \mathrm{rate}} &= \frac{\mathrm{number} \; \mathrm{of} \; \mathrm{success}}{\mathrm{total} \; \mathrm{number} \; \mathrm{of} \; \mathrm{runs}}
\label{eq:sr_eq}
\end{align}

\subsection{Robots and interface setup}

\begin{figure}[tb]
    \centering
    \subfigure[Leap Motion device]{
        \includegraphics[keepaspectratio=true,width=0.24\linewidth]{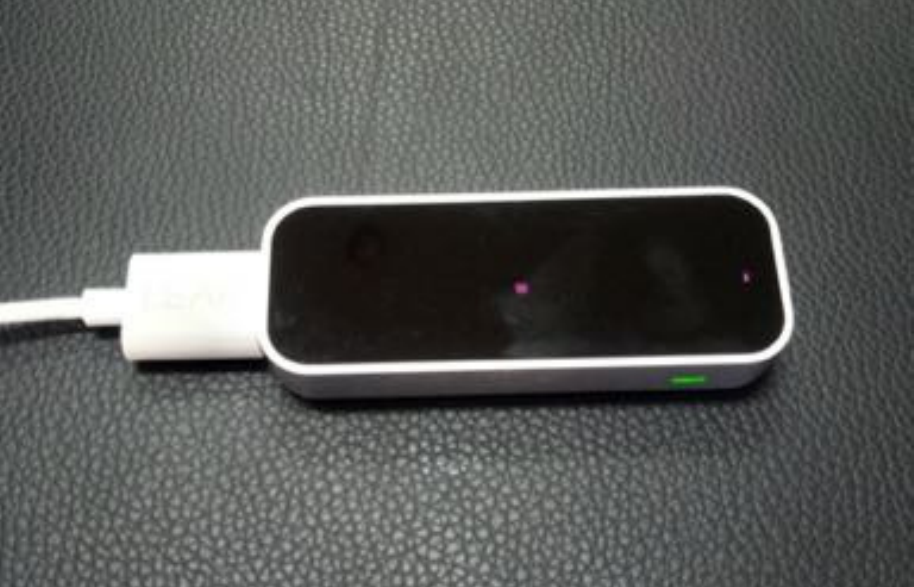}
        \label{fig:lm_device}
    }
    \centering
    \subfigure[Qbchain Yaw-Pitch-Pitch-Pitch (YPPP)]{
        \includegraphics[keepaspectratio=true,width=0.21\linewidth]{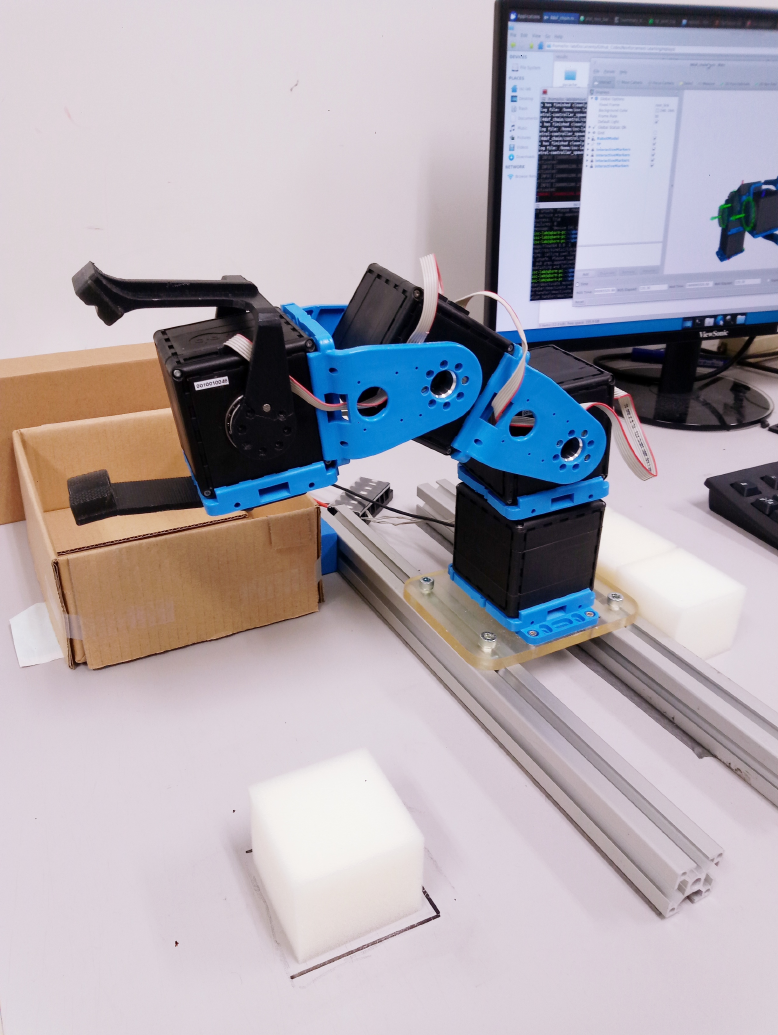}
        \label{fig:yppp}
    }
    \centering
    \subfigure[D'Claw robot]{
        \includegraphics[keepaspectratio=true,width=0.23\linewidth]{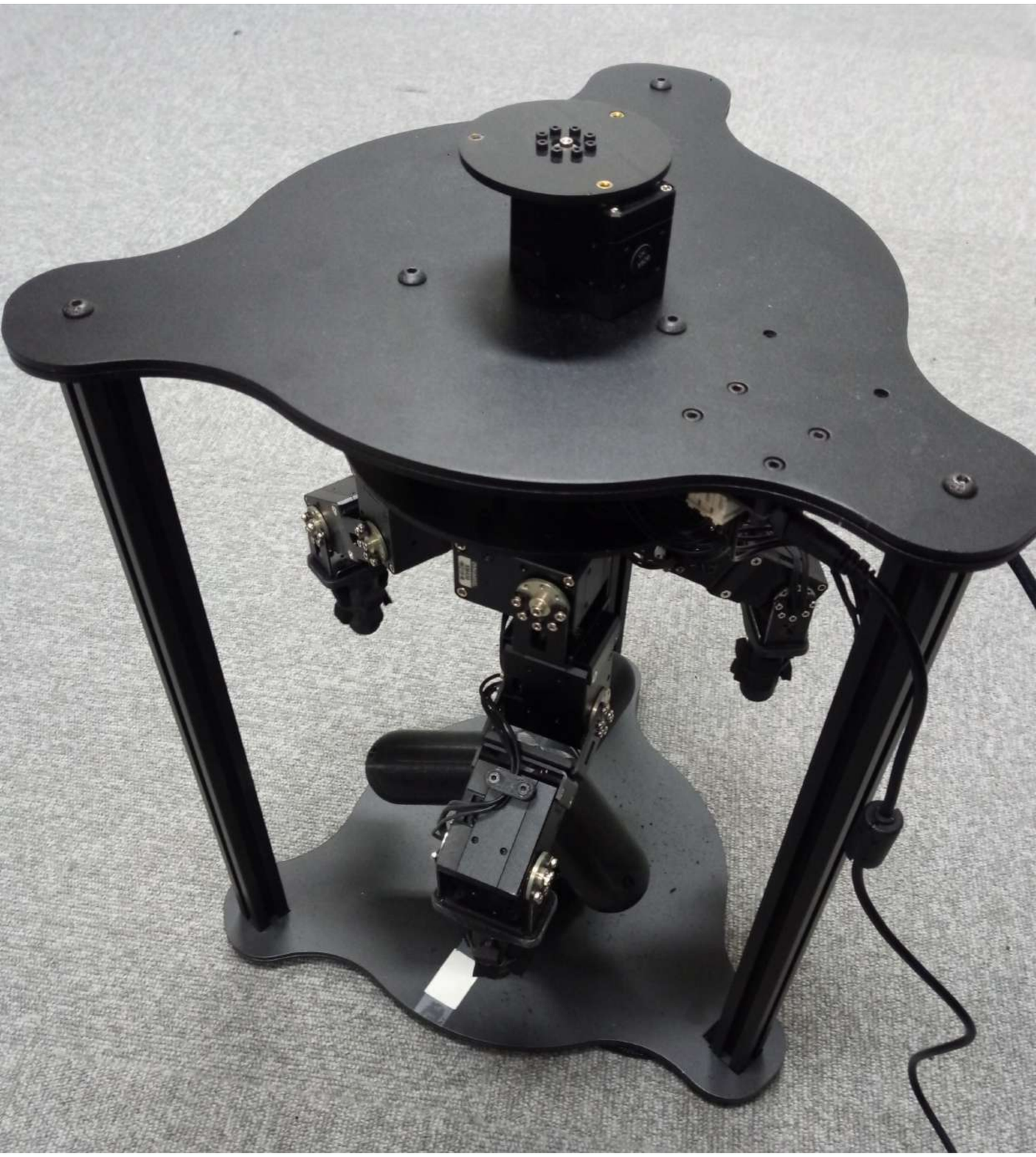}
        \label{fig:dclaw}
    }
    \caption{Robots and interface used in the BC experiments.}
    \label{fig:device}
\end{figure}

\subsubsection{Leap Motion hand tracking device}

 Leap Motion (see Fig.~\ref{fig:lm_device}) is a hand tracking device that captures the movement of the hands and fingers by using optical sensors and an infrared light. The field of view (FOV) of the sensors is about 150 degrees and the detection range goes roughly from 25 to 600 millimeters above the device.
  Each object (arm, hand or finger) detected in the FOV of the device is represented by a program class that encodes various informations such as the position, velocity, direction and other characteristics about the object.

\subsubsection{Qbchain robot and control interface}

 The qbmove~\cite{catalano2011vsa} is a one degree of freedom (1-DoF) modular actuator with a cubic shape of approximately 66 millimeter width. Its stiffness can also be controlled on the hardware level, but is fixed in the following experiments for simplicity. As can be seen in Fig.~\ref{fig:yppp}, the robotic arm employed in this section's experiments is made of 4 cubes assembled such that the first joint axis is vertical, while the three others are horizontal, allowing for an up-and-down and circular motion of the end effector, which consists of a gripper.

 The interface between the Leap Motion device and the qbmove robotic arm developed to allow a human operator to control the robot uses the palm position and grab strength of the Leap Motion's first detected hand.
 The palm position is used as the position of the robot's end effector and an Inverse Kinematics (IK) algorithm is employed to compute the first three joints' angular position. In the experiment, \emph{ikpy} is employed and corresponds to a python inverse kinematics library that can import the kinematic chain of the robot from an URDF file and can quickly approximate the IK solution by employing an iterative optimizer. The obtained joints position values are then sent to the qbchain to move the tip of the fixed part of the gripper. The grab strength is then mapped to the last joint in other to open and close the gripper.

\begin{figure}[tb]
    \centering
    \captionsetup{justification=centering}
    \includegraphics[keepaspectratio=true,width=0.7\linewidth]{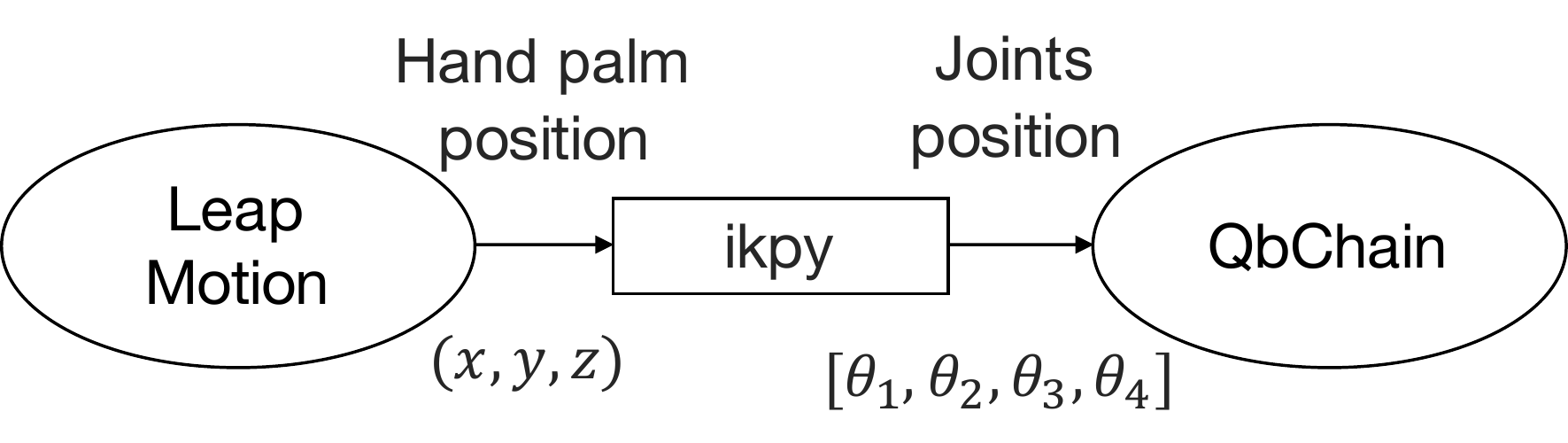}
    \caption{Qbchain-Leap Motion control interface. $[\theta_1, \theta_2, \theta_3, \theta_4]$ are the desired joints angular position.}
    \label{fig:interface}
\end{figure}

 The schematic of the interface is depicted in Fig.~\ref{fig:interface}.

\subsubsection{D'Claw robot and control interface}

D'Claw is a platform introduced by project-ROBEL (RObotics BEnchmarks for Learning)~\cite{ahn2020robel} for studying and benchmarking dexterous manipulation. It's a nine degrees of freedom (DoFs) platform that consists of three identical fingers mounted symmetrically on a base, as shown on Fig.~\ref{fig:dclaw}.

 Its control interface also uses the leap motion device. In particular, the position of the fingers --- the index, the ring and thumb fingers --- of the operator is used to control the three fingers of the robot, again through the \emph{ikpy} library.

\subsection{Qbchain robot experiment}

\subsubsection{Conditions of the experimentation}
 A simple pick-and-drop task is defined, where the goal is to pick an object, here a soft cube, and drop it inside a box, with an observation consisting of a direct state measure containing information about the angle, the angular velocity and the torque (effort), for each of the four joints (hence, the state space dimension equals $12$).
 The action space dimension, on the other hand, is set to be equal to $4$ and corresponds to the desired next angle of the joints (i.e. position controller).

 During training, a Gaussian white noise is added to the states by using a scale factor $\eta = 0.03$, i.e. $s = s + \eta\mathcal{N}(0,1)$, in order to augment the dataset and improve the generalization ability of the models. A small batch size of $32$ is used to reduce the computational cost, and to drive the ability of the gradient updates to escape from local optima.

\subsubsection{Dataset description}
 $90$ trajectories are collected and then divided into $56$ expert trajectories that are almost perfect, and $34$ amateur trajectories that contain hesitant or poor demonstrations.
 The expert trajectories are then further split into two data sets; one, containing $36$ trajectories, for training and another one for validation, comprised of the remaining $20$ trajectories.

\subsubsection{Results}
\begin{figure}[tb]
    \centering
    \captionsetup{justification=centering}
    \includegraphics[keepaspectratio=true,width=0.7\linewidth]{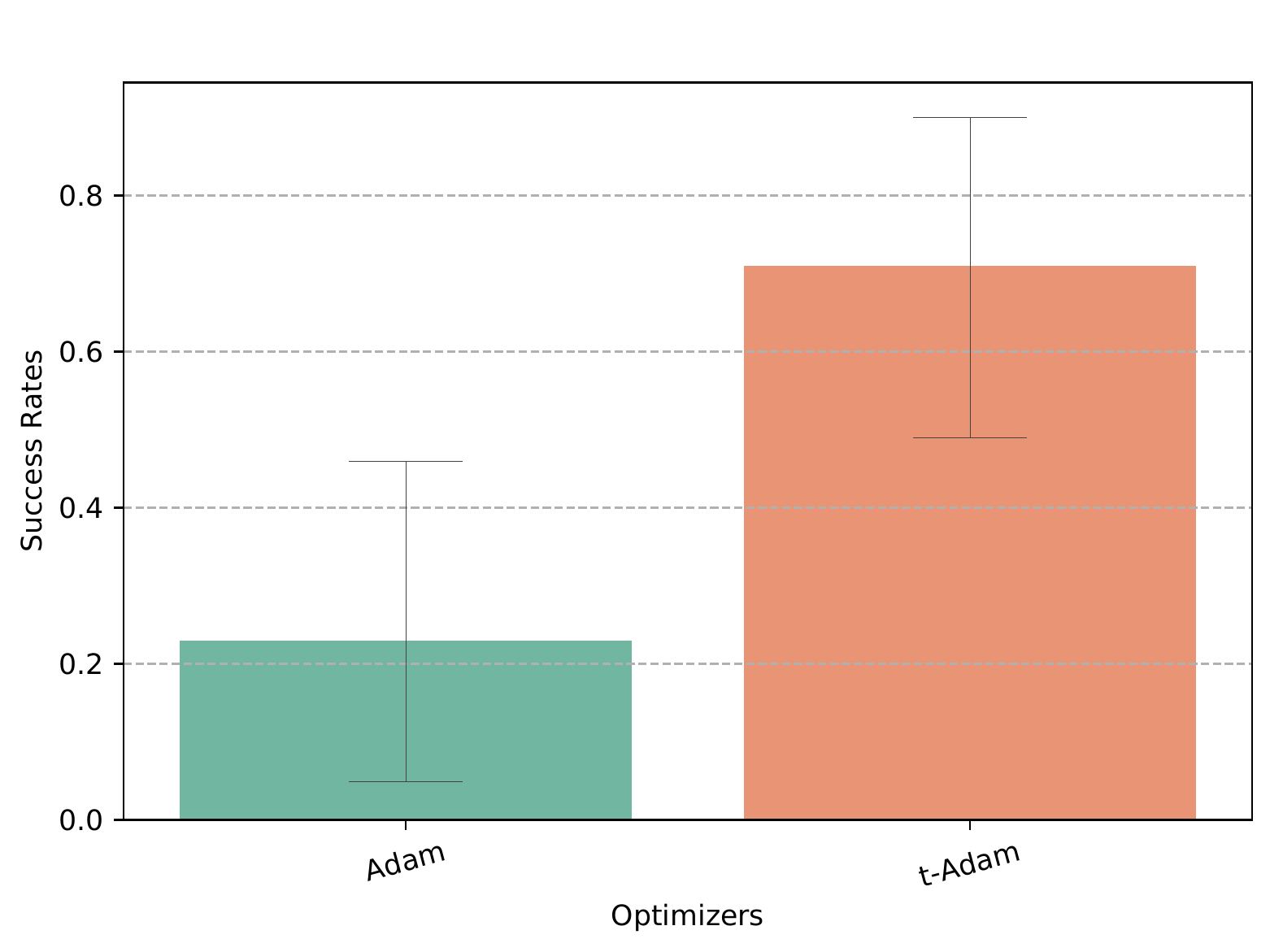}
    \caption{Success rates on the Qbchain robot of the models trained with both amateur and expert demonstrations.}
    \label{fig:sr1}
\end{figure}

 The tests results on the robot, for $10$ trained policies, are given by the success rate over all models and summarized in Fig~\ref{fig:sr1} where the error bars correspond to the $95\%$ confidence interval. This success rate is computed by running each trained model $10$ times (i.e. total number of runs = $10$) and Eq.~\eqref{eq:sr_eq} is employed by counting the number of times the model is able to solve the task (i.e. pick the object and drop it in the box). Each episode is ran with a fixed budget of $40$ steps and a model is said to have failed if it is not able to complete the task within this number of steps.

The success rates in Fig~\ref{fig:sr1} show that, using a robust optimization method such as the t-momentum based Adam algorithm, it is possible to efficiently train a behavioral cloning agent with datasets that contain not only expert demonstrations, but also amateur performances.

\begin{figure}[tb]
    \centering
    \captionsetup{justification=centering}
    \includegraphics[keepaspectratio=true,width=0.75\linewidth]{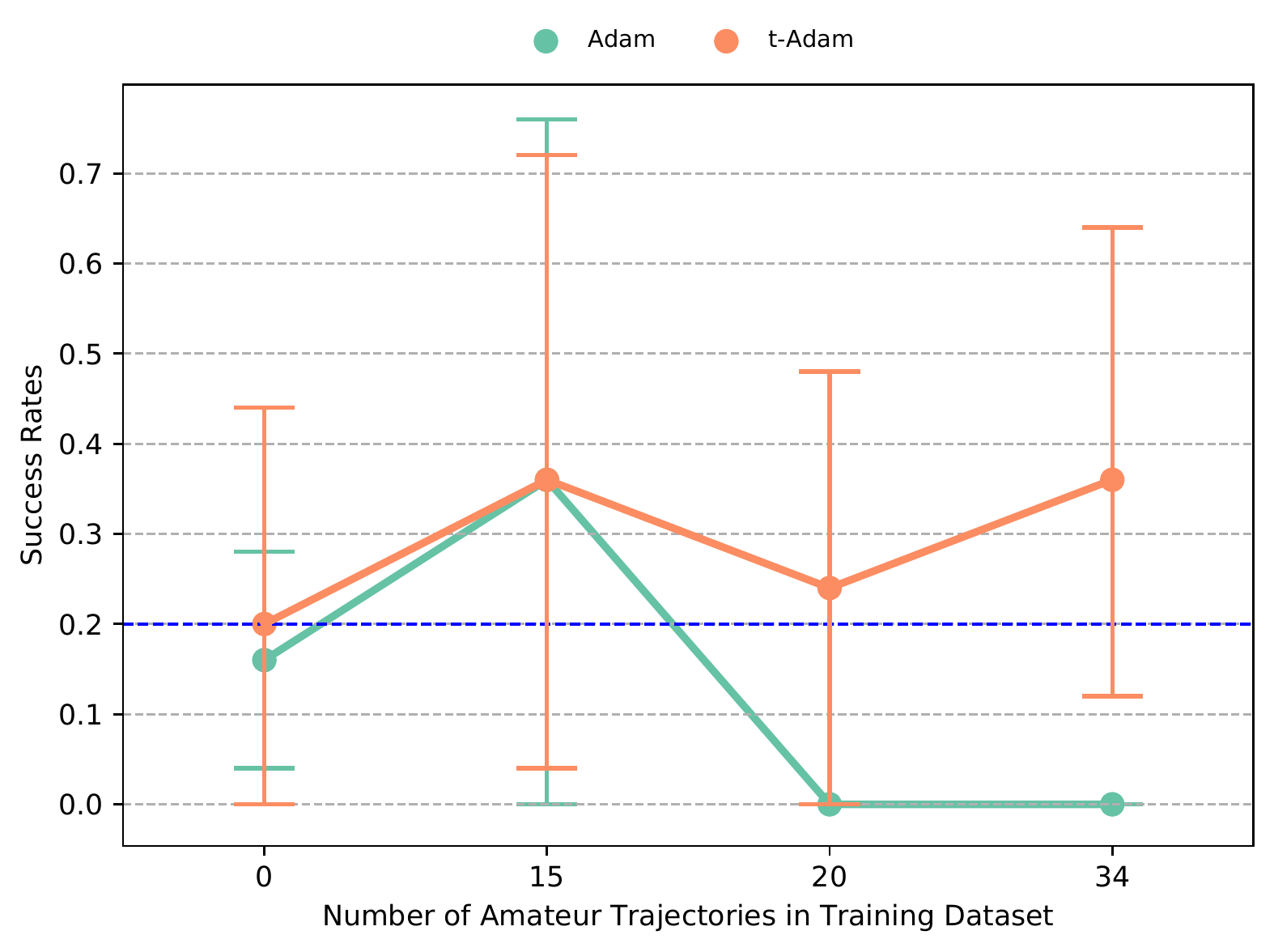}
    \caption{How the amateur data can be useful: Success rates of the trained models on the Qbchain robot with various amateur data proportion.}
    \label{fig:sr2_amateurAdvantage}
\end{figure}

 Fig.~\ref{fig:sr2_amateurAdvantage}, where the success rate of $5$ trained models is summarized with $5$ total number of runs per model, displays the contribution of the amateur demonstrations. Indeed, we can see that, when considering a small number of expert demonstrations (i.e. $15$ trajectories), the addition of the demonstrations containing imperfect $(s, a)$ pairs increases the success rate of the models trained with the robust t-momentum optimizer. This result highlights the fact that amateur demonstrations are useful and can be used to augment the size of the training dataset, instead of being discarded as it is usually done in BC.

\begin{figure}[tb]
    \centering
    \captionsetup{justification=centering}
    \includegraphics[keepaspectratio=true,width=0.75\linewidth]{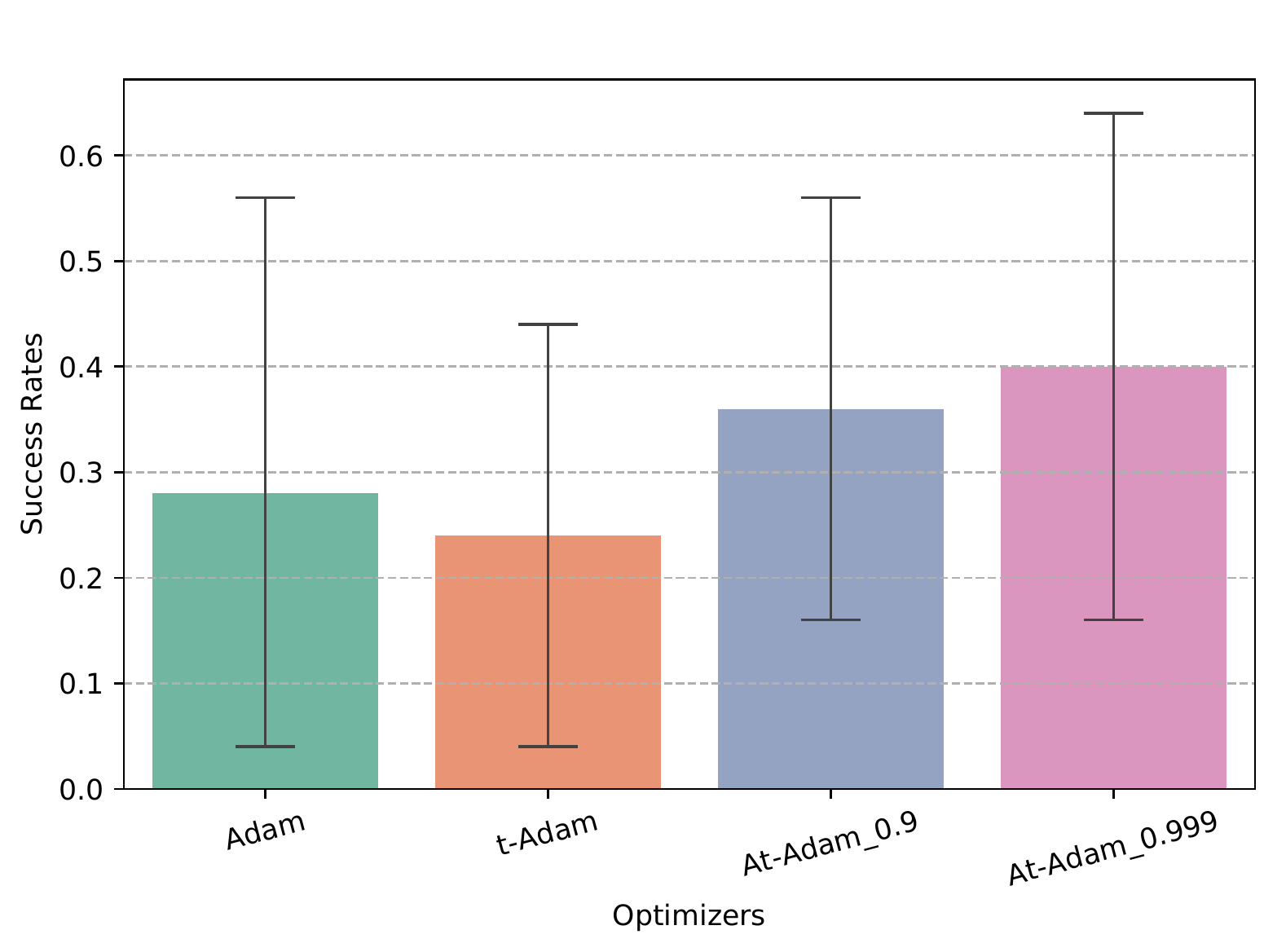}
    \caption{At-Adam Advantage: Success rates on the Qbchain robot of the models trained without noise and amateur data.}
    \label{fig:sr3_rbc_atadam}
\end{figure}

 However, in Fig.~\ref{fig:sr3_rbc_atadam}, after removing the amateur demonstrations and setting the noise scale factor to $\eta = 0$, we computed the success rates by running again $5$ trained models $5$ times each (i.e. total number of runs = $5$). With this modification, we can see that in the absence of imperfect demonstrations and without the Gaussian noise for state augmentation, the Adam optimizer performs better than t-Adam, due to the fixed high robustness of the later.

\begin{figure}[tb]
    \centering
    \captionsetup{justification=centering}
    \includegraphics[keepaspectratio=true,width=0.75\linewidth]{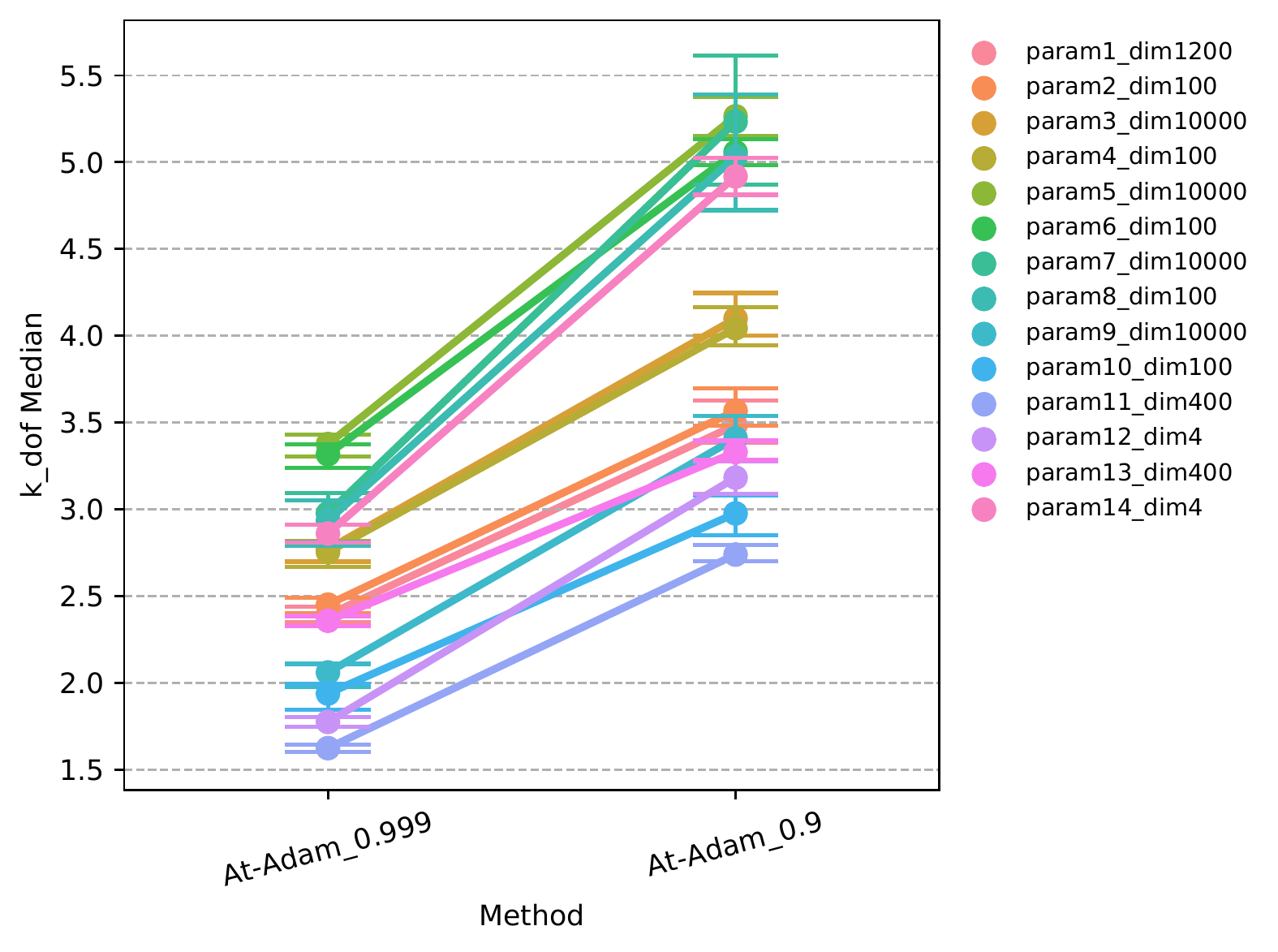}
    \caption{At-Adam: median of the adapting degrees of freedom's factor $k$ during the learning for each parameter of the network.}
    \label{fig:kdofs_atadam}
\end{figure}

 This result allows us to display the importance of the adaptive robustness feature of At-Adam. Indeed, in the same Fig.~\ref{fig:sr3_rbc_atadam}, we see how the adaptive t-momentum optimizer improves the success rates of the imitators and performs even better than Adam. Hence, the adaptive robustness unarguably allows it to extract more optimal information from the expert dataset than what is allowed with non-robust methods. At-Adam, thanks to its automatic robustness adjustment, is able to find a compromise between the too-robust t-Adam with its $k = 1$ and the non-robust Adam with its $k = \infty$, outperforming both methods. Fig.~\ref{fig:kdofs_atadam} shows the median of the adapting degrees of freedom's factor $k$ during the learning. We can see that At-Adam has a median robustness parameter higher than $1$.

\subsection{D'Claw robot experiment}

 To further confirm the ability and limitation of the robust BC with the adaptive t-momentum algorithm to adapt to different ratio of imperfect demonstrations, we conducted the following experiments using the D'Claw robot.

\subsubsection{Conditions of the experimentation}
 In the experiments, we define the task to consist in rotating a passive DoF (the object located on the middle of the base in Fig.~\ref{fig:dclaw}) to a fixed target angle. Specifically, the task consists in turning the object from the angle $0.0$ to the target angle $\pi$, with the success being achieved if the object's position falls within the range $(\pi-0.1, \pi+0.1)$. The state space is given by the angular position and velocity of the fingers' nine joints, the target position and the current angular position of the object along with their cosine and sine values, the object's velocity and finally a success flag and the error between the current position and the target position, for a total dimension of $27$. The actions' dimension is set to $9$ corresponding to the position of the fingers' joints.
 The batch size is again set to $32$, but this time no noise is included in the states during training.

\subsubsection{Dataset description}
 For this task, only 34 demonstrations are recorded, consisting in 14 amateur demonstrations with imperfect state-action pairs, and 20 expert demonstrations. The expert data is then split in half; one half is used for training and the other half for validation. All the demonstrations were successful ones, where the operator was able to solve the task.


\subsubsection{Results}
\begin{figure}[tb]
    \centering
    \captionsetup{justification=centering}
    \includegraphics[keepaspectratio=true,width=0.75\linewidth]{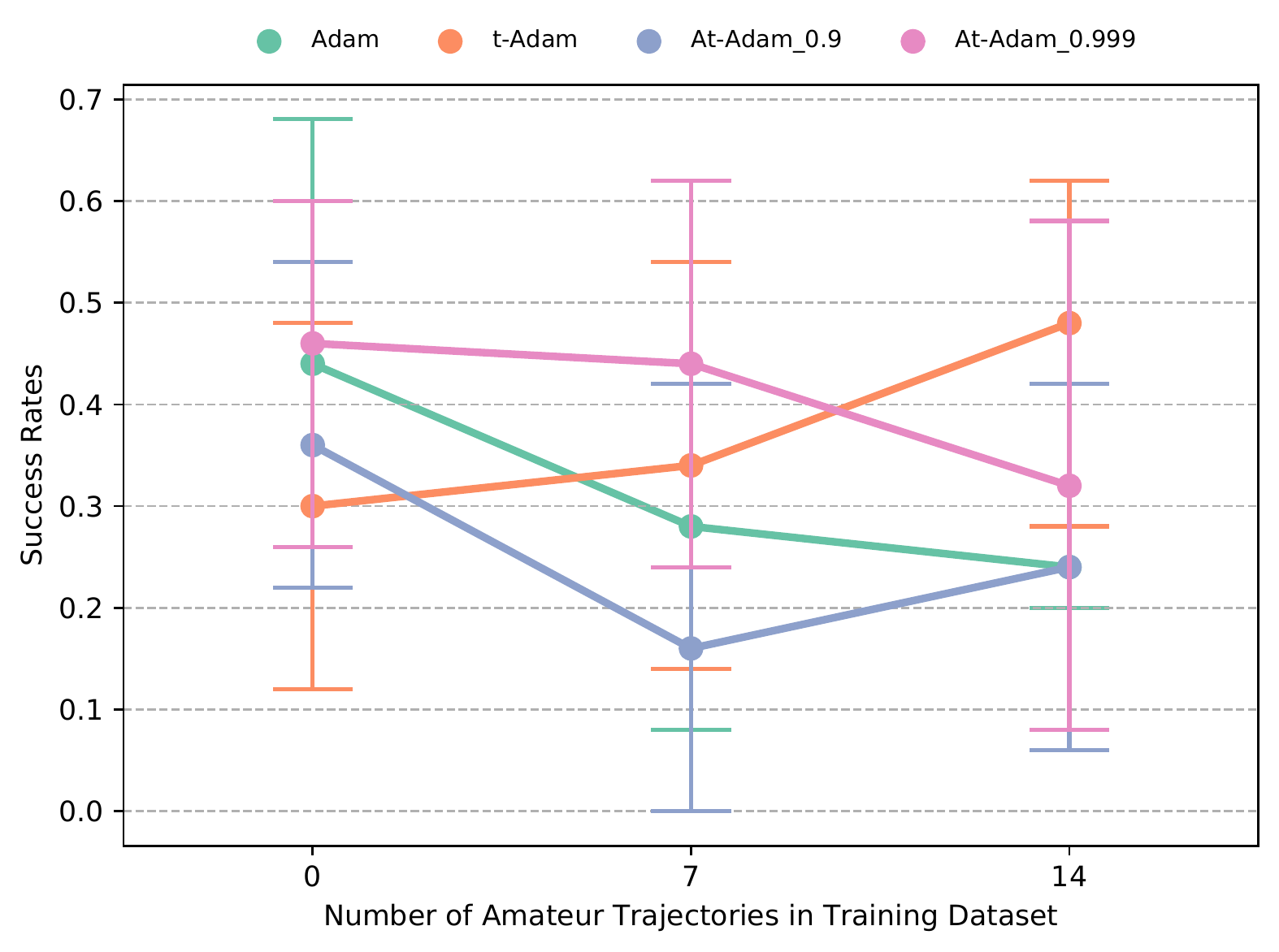}
    \caption{Success rates of the trained models on the D'Claw robot with varying amateur data proportion.}
    \label{fig:dclaw_sr1}
\end{figure}

 Fig.~\ref{fig:dclaw_sr1} shows the average performance of $5$ models with $10$ runs each. Each run is given a fixed budget of $200$ steps and the success is achieved if the imitator is capable of bringing the object's position within the range of the target position, i.e. $\pi \pm 0.1$.
 The success rate of Adam is as expected with the addition of imperfect demonstrations, but the one of At-Adam with $\lambda=0.9$ also suffered a significant decrease.
 On the other hand, At-Adam with $\lambda=0.999$ maintains its performance for half of the amateur demonstrations, but then deteriorates when $14$ amateur trajectories are given.
 Since the success rate of t-Adam with its robustness fixed at $\nu=1$ increased by adding the amateur trajectories, it is likely that the proposed adjustment rule for the t-momentum's degrees of freedom $\nu$ was incomplete, or that the simultaneous optimization of $\theta$ and $\nu$ caused the policy to fall into one of the local solutions when updating $\theta$ with temporarily high $\nu$.

\begin{figure}[tb]
    \centering
    \captionsetup{justification=centering}
    \includegraphics[keepaspectratio=true,width=0.75\linewidth]{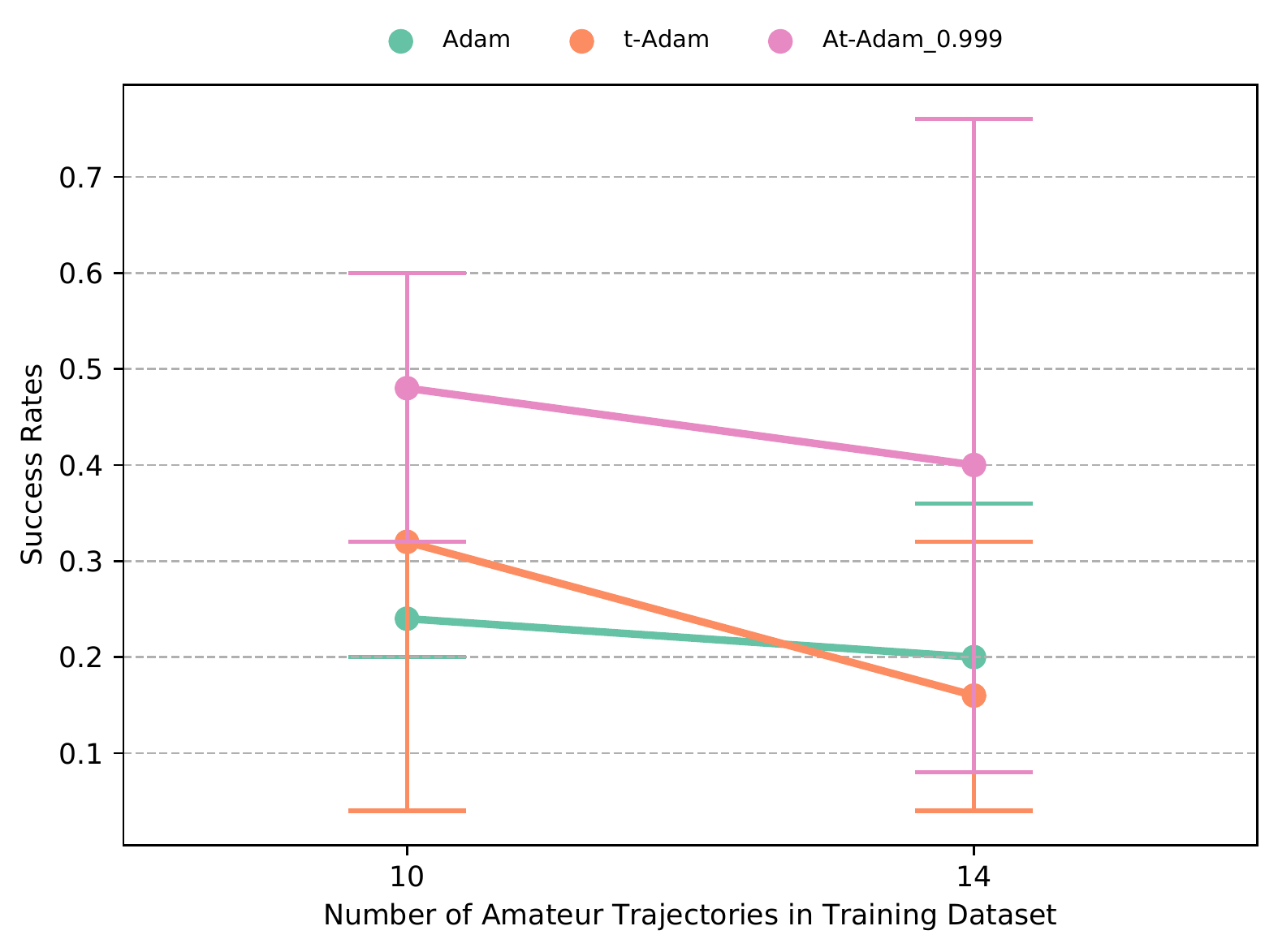}
    \caption{Success rates of the trained models on the D'Claw robot with only amateur data and no expert data.}
    \label{fig:dclaw_sr2}
\end{figure}

 For further investigation, Fig.~\ref{fig:dclaw_sr2} shows the success rates of the models trained using only the amateur data. As we can see, despite being previously affected by the presence of imperfect demonstrations in the previous result, At-Adam is capable of altering its robustness to extract the most useful information from this imperfect dataset.
 Interestingly, with $14$ amateur trajectories, the success rate in Fig.~\ref{fig:dclaw_sr2} is higher than that in Fig.~\ref{fig:dclaw_sr1}. This suggests that the decrease in success rate of At-Adam may be due to a cause outside the proposed method. That is, BC is poor at learning multimodal policies~\cite{ghasemipour2020divergence}, and if the policy optimized by the amateur demonstrations and the one by the expert's are different but both can solve the task, learning with both demonstrations will fail due to the nature of BC.

\section{CONCLUSIONS}

 In this study, we showed how the t-momentum could be used to produce robust imitators under the BC framework. Taking advantage of the Aeschliman's algorithm~\cite{aeschliman2010novel}, we introduced a mechanism to automatically adjust the robustness of the t-momentum strategy, in order to deal with different proportion of imperfect and noisy $(s, a)$ pairs in the demonstrations. The application on two different robots with different tasks having different degrees of difficulties displayed the effectiveness of the proposed approach.

 As implied by the experiments, the amateur demonstrations may make the policy multimodal, hence, this reaffirms the fact that the standard BC and/or the policy model should be modified in order to resolve this multimodality. In addition, the proposed method can be regarded as a kind of safety net, because it removes outliers at the final stage of optimization. An unsupervised classification of demonstrations and/or a robust design of the loss function would be required to actively utilize amateur demonstrations and further bring forth their potential for wide and unlimited imitation learning applications. In future works, the proposed method will be integrated to such algorithms.




%
%
%
%

\bibliographystyle{IEEEtran}
\bibliography{IEEEabrv,bibliography}

\end{document}